\pdfoutput=1

\documentclass[11pt]{article}

\usepackage{acl}

\usepackage{times}
\usepackage{latexsym}

\usepackage[T1]{fontenc}

\usepackage[utf8]{inputenc}

\usepackage{microtype}

\usepackage{tikz}
\usetikzlibrary{positioning, shapes.geometric}
\usepackage{bbding}
\usepackage{times}
\usepackage[toc,page]{appendix}
\usepackage{latexsym}
\usepackage{natbib}
\usepackage{graphicx}
\usepackage{subfigure}
\usepackage{amsmath}
\usepackage{amsthm}
\usepackage{amsfonts,amssymb}
\usepackage{booktabs}
\usepackage{multicol}
\usepackage{multirow}
\usepackage{diagbox}
\usepackage{ulem}
\usepackage{bm}
\usepackage{pifont}

\usepackage{cleveref}
\crefname{section}{§}{§§}
\Crefname{section}{§}{§§}
\usepackage{dashrule}
\newtheorem{innercustomgeneric}{\customgenericname}
\providecommand{\customgenericname}{}
\newcommand{\newcustomtheorem}[2]{%
  \newenvironment{#1}[1]
  {%
   \renewcommand\customgenericname{#2}%
   \renewcommand\theinnercustomgeneric{##1}%
   \innercustomgeneric
  }
  {\endinnercustomgeneric}
}
\usepackage{caption}

\newcustomtheorem{customthm}{Theorem}
\newcustomtheorem{customcoro}{Corollary}

\usepackage[linesnumbered,boxed,ruled,commentsnumbered]{algorithm2e}
\usepackage{lipsum}

\newcommand{\STAB}[1]{\begin{tabular}{@{}c@{}}#1\end{tabular}}

\title{A Semi-Autoregressive Graph Generative Model \\ for Dependency Graph Parsing}

\author{Ye Ma, Mingming Sun, Ping Li \\ 
        Cognitive Computing Lab\\
        Baidu Research \\ 
No.10 Xibeiwang East Road, Beijing 100193, China\\
10900 NE 8th St. Bellevue, Washington 98004, USA\\
        \texttt{\{maye811906,sunmingming01, pingli98\}@gmail.com}}

\begin{document}
\maketitle
\begin{abstract}
Recent years have witnessed the impressive progress in Neural Dependency Parsing. According to the different factorization approaches to the graph joint probabilities, existing parsers can be roughly divided into autoregressive and non-autoregressive patterns. The former means that the graph should be factorized into multiple sequentially dependent components, then it can be built up component by component. And the latter assumes these components to be independent so that they can be outputted in a one-shot manner. However, when treating the directed edge as an explicit dependency relationship, we discover that there is a mixture of independent and interdependent components in the dependency graph, signifying that 
both aforementioned models fail to precisely capture the explicit dependencies among nodes and edges. Based on this property, we design a Semi-Autoregressive Dependency Parser to generate dependency graphs via adding node groups and edge groups autoregressively while pouring out all group elements in parallel. 
The model gains a trade-off between non-autoregression and autoregression, which respectively suffer from the lack of target inter-dependencies and the uncertainty of graph generation orders. The experiments show the proposed parser outperforms strong baselines on Enhanced Universal Dependencies of multiple languages, especially achieving $4\%$ average promotion at graph-level accuracy. Also, the performances of model variations show the importance of specific parts.

\end{abstract}


\section{Introduction}\label{sec:intro}
Dependency graph in neural parsing is a directed graph representing semantic dependencies between words, with a transitive relation traveling from the rooted node to all words in the sentence phase by phase. As such, transition-based parsing seems to be a natural choice, as it builds up the parsing graph sequentially so that the dependency relationships can be captured. However, graph-based parsing dominates recent competitions on parsing technologies including IWPT 2020 and 2021 \cite{bouma2020overview, bouma2021raw}, even if using a simple biaffine attention \cite{dozat2017deep} only to predict the whole graph at once.
To explore a more effective parsing method that can represent these dependency relationships in a rigorous manner, we define and construct Topological Hierarchies for dependency graphs based on the explicit dependencies carried by them. According to the characteristics of topological hierarchies, we
 proposes a {\bf S}emi-{\bf A}utoregressive Dependency {\bf G}raph Pars{\bf er} ({\bf SAGER}) -- a novel graph-based parsing fashion via the semi-autoregressive graph generation.

\begin{figure}[t]
\centering
\includegraphics[width=0.49\textwidth]{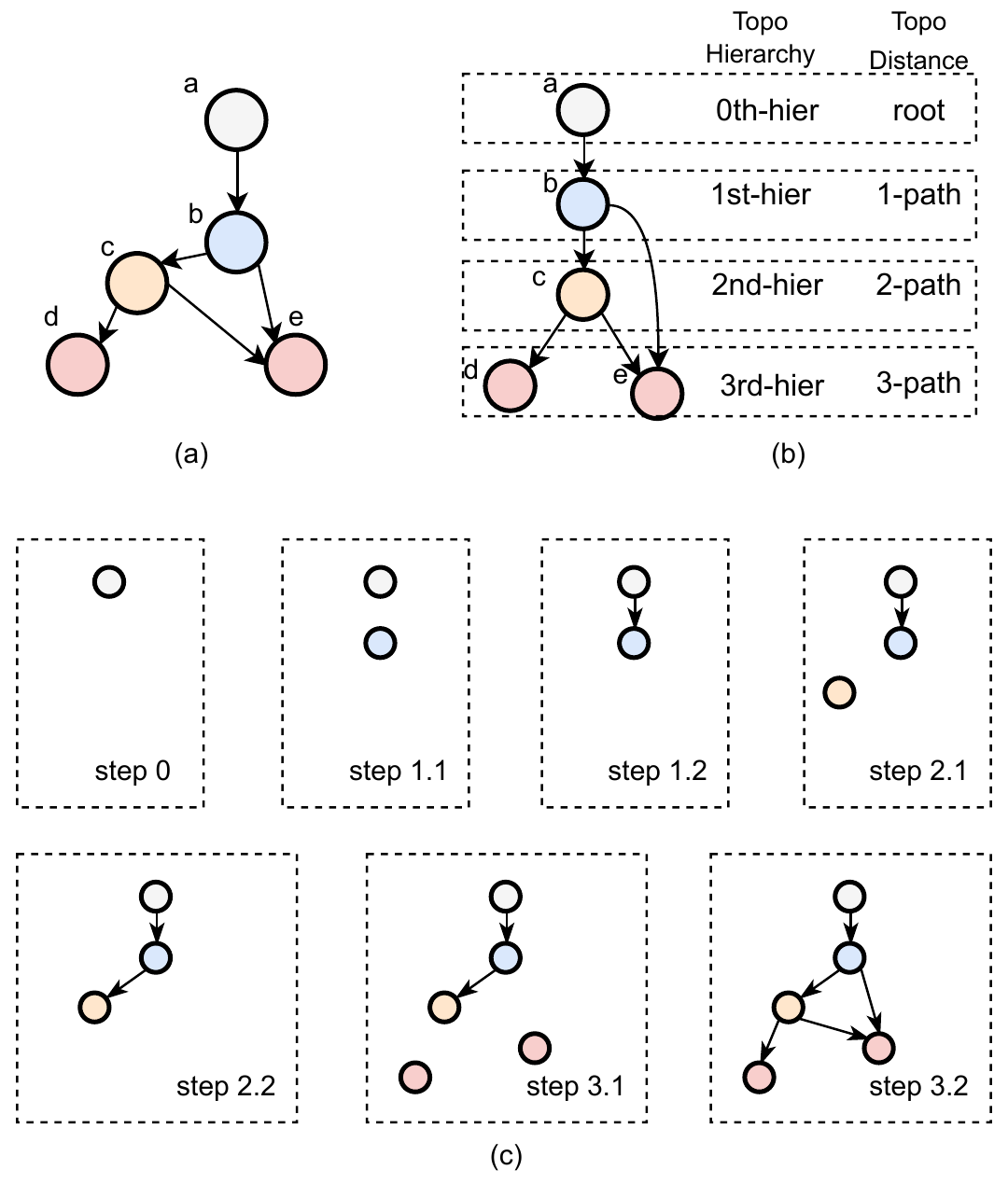}

\vspace{-0.1in}

\caption{\label{fig:intro} (a) An example graph (b) Divide nodes into different topological hierarchies based on their furthest distances from the root node. (c) Semi-autoregressive graph generation process.}\vspace{-0.05in}
\end{figure}

Generally, autoregressive graph generation indicates that the model dynamically adds nodes and edges based on the generated sub-graph structure until reaching the complete graph. It brings a challenge to determine the generation order of graphic data, because there is no conventional reading order like text data \cite{chen2021order}.
Although the dependency graphs stipulate the strict sequential dependencies by directed edges, there is a lack of such topological orders between sibling nodes. For instance, node $\mathbf{b}$ in Figure \ref{fig:intro}.a depends on the node $\mathbf{a}$ because there is an explicit edge pointing from $\mathbf{a}$ to $\mathbf{b}$. However, it is hard to decide the dependency relationship between node $\mathbf{d}$ and node $\mathbf{e}$ as they are not linked neither directly nor indirectly. Previous works on directed graph generation solve the problem surfacely. \citet{cai2019core, cai2020amr} sort these sibling nodes randomly at the early stage of training and then change them to a deterministic order (e.g., relation-frequency) at later training steps. Some other works do the sorting by referring to the orders of the known sequences like word order or alphanumerical order \cite{zhang2019amr, zhang2019broad, bevilacqua2021one}. Such imposed orders would bring more exposure bias \cite{ranzato2016sequence} to teacher forcing training. More specifically, these sibling nodes are essentially orderless and non-interlogical. They cannot form stationary and logical patterns like words in sentences. As a result, it is hard for the sibling nodes to be generated in the same order as in the training, causing learned knowledge invalid and later predictions misled. 
The aforementioned random ordering seems to alleviate the problem to some extent, but it destabilizes and complicates the training process and generally results in inferior models.

Instead of imposing orders on these sibling nodes, we assume them (including their incoming edges) to be conditionally independent to construct topological hierarchies as the generation orders. As shown in Figure \ref{fig:intro}.b, we divide nodes into several hierarchies according to their furthest distances from the root node. We can see that there are no explicit dependency relationships between nodes in the same hierarchy. Besides, nodes in the later hierarchies only depends on those in the previous hierarchies, forming a  natural generation sequence. For a directed acyclic graph (DAG), it {\it at least} has one topological ordering but {\it only} has one topological hierarchy. At each generation step, we firstly predict all new nodes in parallel and then calculate their incoming edges by the biaffine attention \cite{dozat2017deep}. In a word, our model autoregressively adds node groups and edge groups but non-autoregressively generates elements in these groups. See Figure \ref{fig:intro}.c for our semi-autoregressive generation process.

Another challenge lies at the probable edge sparsity problem when representing sub-graph structures. Sparse graph representation models like GCN \cite{kipf2017semi} heavily rely on the given adjacency to capture context information. That means it may fail to represent historical information completely and efficiently when predicted edges make mistakes. An extreme situation of edge sparsity is that the new nodes have no incoming edges predicted so that the model can only represent its node features rather than the sub-graph structure.
To enhance the robustness of the generator, we adopt the dense structure of Transformer \cite{vaswani2017attention}. In our Graph Transformer, there are implicit edges linking from the nodes in the previous and current hierarchies to the new node.
Then, the predicted explicit edges serve as the bias to adjust the attention distribution over the implicit edges so that the model can adaptively select useful structural information.

Overall, this paper proposes a novel direction -- semi-autoregression to deal with dependency parsing problems, distinguished with autoregression and non-autoregression (detailed definitions about them are available in \cref{sec:related}). With the dependencies denoted as the directed edges, the semi-autoregressive pattern unfolds graphs in the order of topological hierarchies, which strictly follows the explicit dependency relationships defined in dependency graphs. 
Our model strikes a reasonable balance, resulting in it neither ignoring inter-output dependencies like non-autoregressive models, nor suffering from severe exposure bias in generation orders like autoregressive models
On the experimental side, we evaluate SAGER on Enhanced Universal Dependencies (EUD) which are non-tree dependency graphs. In addition to the official evaluation metric Enhanced Label Attachment Scores (ELAS), we design a graph-level matching score (GMS) to assess the probability of returning an completely correct graph.  Our model outperforms other baselines in both metrics especially GMS, showing that it is good at overcoming some intractable prediction errors.
Finally, we introduce multiple model variations to investigate the effect of different model components and show that our model is well-designed, especially the parts of discarding imposed orders and adding implicit edges.

\section{Related Work}\label{sec:related}

{\bf Autoregressive Parser}. Generally, a generator is in an autoregressive fashion provided its generation probability at each step is conditional on items it produces previously. Transition-based parser obviously conforms to the characteristic, as it updates the action probability every step based on the words, tags and label embeddings previously put in the buffer and stack \cite{chen2014fast}. Meanwhile, we note that some mechanisms commonly used in autoregressive generators are used to improve transition-based parsers like beam search and pointer networks \cite{weiss2015structured, ma2018stack,fernandez2019left}. On the other hand, \citet{cheng2016bi} proposes a graph-based autoregressive parser by adding arcs sequentially with the considerations of previous parsing decisions. However, it should not be taken as a rigorous graph generative model as it does not generate by extending the sub-graph structures. Actually, instead of dependency graphs, it is more prevalent that leverage the autoregressive graph generators to parse Abstract Meaning Representation (AMR) \cite{cai2019core, cai2020amr, zhang2019broad, zhang2019amr}. They are all in the (fully) autoregressive pattern that an order is imposed to nodes and edges without topological orderings. In this paper, we focus on investigating the effects of these imposed orderings by introducing some variations of the proposed model. 

\vspace{0.1in}
\noindent{\bf Non-Autoregressive Parser}. In contrast, non-autoregression implies that all components factorized from the graph are independent, so their probabilities do not affect each other and can be obtained in parallel at any time. A representative non-autoregressive parser is Deep Biaffine Attention (BiAtt) \cite{dozat2017deep} which assuming all edges are independent. For the tree-structure dependency graphs, it is often followed by a searching algorithm for the Maximum Spanning Tree (MST). Some heuristic algorithms \cite{li2020global,kiperwasser2016easy} construct the MST step by step, which yet does not mean they are in the autoregressive manner because all edge probabilities are predicted at once and fixed before the searching. As to the higher-order graph-based parsers, \citet{ji2019graph} incorporates the second-order knowledge into the word representations and still uses the BiAtt as the final parser. \citet{wang2019second,zhang2020efficient} decompose the graph into components of different second-order parts. Different from BiAtt that each component is an edge, here some components consists of two edges whose joint probabilities can be calculated as a whole by a trilinear function. They still belong to non-autoregressive parsers because their components are independent of each other and disable to be subdivided.

\section{Proposed Model}
\subsection{Definitions}

\vspace{0.1in}
{\bf Problem Definition}. Conditional on the source sentence $S =(w_n)_{n=1}^N$, the task is to generate a dependency graph hierarchy by hierarchy. The generation process can be denoted as a sequence of components: $(C^{(t)})_{t=0}^T, T \leq N$. We firstly turn dependency graphs to DAGs by deleting the back edges in their cycles. It should be mentioned that there are only a few graphs with cycles and we can add these removed edges back by rules before evaluations. Then we can construct topological hierarchies based on the furthest distance from each node to the root node. The initial component $C^{(0)}=\{v_0\}$ in the $0$-th hierarchy only has a root node. When $t > 0$, the component in the $t$-th hierarchy is defined as $C^{(t)} = \{V^{(t)}, E^{(t)} \}$.
Let $\mathcal{V}_t = \bigcup_{j=0}^{t} V^{(j)}$, then $V^{(t)} = \{v_i\}_{i=|\mathcal{V}_{t-1}|}^{|\mathcal{V}_t|-1} $
is a set of nodes in the $t$-th hierarchy. And, $E^{(t)} = \{(v_j, v_i, z_{ji})|v_j\in \mathcal{V}_{t-1}, v_i \in V^{(t)}
\}$ is a set of edges pointing from nodes in the previous hierarchies to the current nodes, where $v_j$ is the head of $v_i$ and $z_{ji}$ is the label on the arc.

\vspace{0.1in}
\noindent{\bf Explicit and Implicit Edge}.
We define two kinds of edges, namely {\it explicit} edges and {\it implicit} edges. The former is what we need to really predict. Let $\mathcal{N}_i$ be the explicit first-order neighbours of the node $v_i \in V^{(t)}$ and $\mathcal{D}_i$ be the implicit neighbours, and $\mathcal{N}_i \cup \mathcal{D}_i = \mathcal{V}_t$. Notably, nodes in $\mathcal{N}_i$ can not appear in $V^{(t)}$ according to the definition of topological hierarchy. They have uni-directional edges pointing to the node $v_i$ with arc labels, and these edge can be found in $E^{(t)}$. On the other hand, nodes in $\mathcal{D}_i$ should not have pointed to $v_i$, but our model does so because we expect nodes to learn structural information adaptively. It should be mentioned that nodes in the same component or hierarchy also have implicit edges linking to each other, i.e., $V^{(t)} \subseteq \mathcal{D}_i$.

\begin{figure*}
\centering
\includegraphics[width=0.9\textwidth]{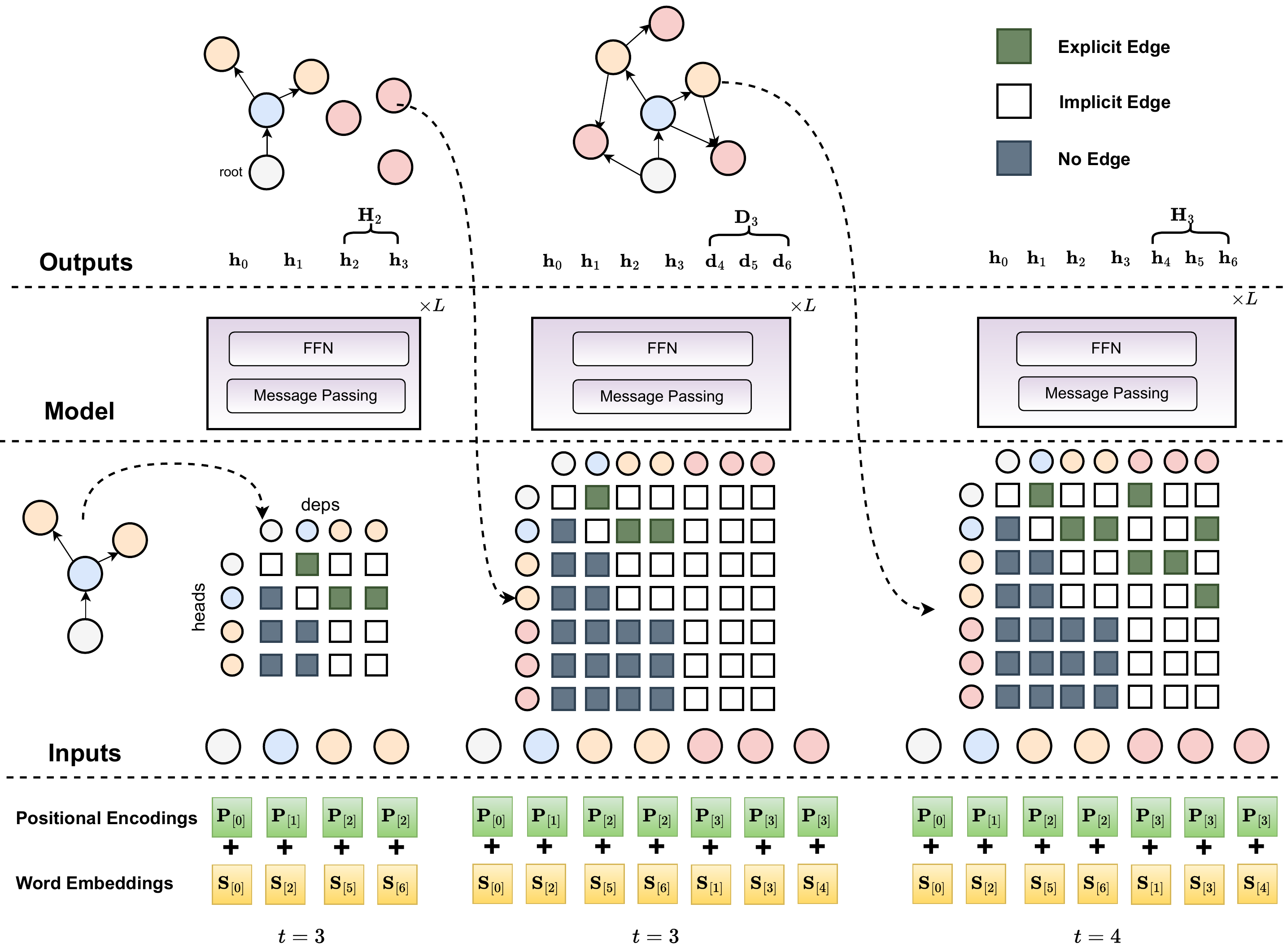}

\caption{\label{fig:flow} Semi-autoregressive generation process and graph transformer. }
\end{figure*}

\newpage

\noindent{\bf Head and Dependent Representation}.
We define two representations of the same node with different roles, namely the {\it head} representation and the {\it dependent} representation. Each generated node will first be used as a dependent node to calculate its incoming arcs, and then as a head node until the end of generation. 
We define the head vector of a node $v_i \in V^{(t)}$ as $\mathbf{h}_i$ and its dependent vector as $\mathbf{d}_i$. For a component, its head matrix $\textbf{H}^{(t)} = \mathcal{F}_{\theta} (\mathcal{V}_t,\mathcal{E}_{t}, S) $ and dependent matrix $\textbf{D}^{(t)} =\mathcal{F}_{\theta} (\mathcal{V}_t,\mathcal{E}_{t-1}, S)$ are the concatenations of multiple corresponding node representation, where $\mathcal{E}_{t} = \bigcup_{j=0}^t E^{(t)}$. 
We can see that the difference between them is that the latter inputs lack $E^{(t)}$, which means there are no available explicit edges pointing to $V^{(t)}$ nodes when calculating dependent representations.  We shall mention that the graph representation model $\mathcal{F}_{\theta}(\cdot)$ can represent all components, but we only need to focus on the new component at each generation step because the new component does not affect node representations in the previous components.

\vspace{0.1in}
\noindent{\bf Training Objective}. The objective is to maximize the graph joint probability $\mathcal{J}$:
\begin{equation}
    \mathcal{J}=\prod_{t=1}^{T} P(V^{(t)}|\mathcal{V}_{t-1}, \mathcal{E}_{t-1}) P(E^{(t)} | \mathcal{V}_{t}, \mathcal{E}_{t-1})\\\notag
\end{equation}
\begin{equation}
     P(V^{(t)}|\mathcal{V}_{t-1}, \mathcal{E}_{t-1}) = \prod_{v_i \in V^{(t)}} P(v_i|\mathcal{V}_{t-1}, \mathcal{E}_{t-1}) \\\notag
\end{equation}
\begin{equation}
    P(E^{(t)} | \mathcal{V}_{t}, \mathcal{E}_{t-1})  = \prod_{e_i \in E^{(t)}} P(e_i| \mathcal{V}_{t}, \mathcal{E}_{t-1}) \\\notag
\end{equation}
\begin{equation}
    \mathcal{V}_{t} = \mathcal{V}_{t-1} \cup V^{(t)}, \  \mathcal{E}_{t} = \mathcal{E}_{t-1} \cup E^{(t)}\\\notag
\end{equation}

It indicates that we autoregressively generate the new node group $V^{(t)}$ and the edge group $E^{(t)}$ based on groups generated previously and the elements in the same group are independent. 

\subsection{Graph Generation Process}
Figure \ref{fig:flow} presents the generative process from the $3$-rd step to the $4$-th step. Specifically, 
At the generation step $t$, we firstly update head representations $\textbf{H}^{(t-1)}$ for the last-step nodes $V^{(t-1)}$ using their network structure information $E^{(t-1)}$. Notably, although there only generates an intermediate sub-graph of the entire structure, the explicit topological information of $V^{(t-1)}$ nodes is completed because they would not have incoming arcs from nodes generated later. On the other hand, these sentence words have been represented as a dense matrix $\textbf{S}$ by a Transformer-encoder. Then, their probabilities of being selected are calculated by:
\begin{align}\notag
    &P(w_{1:N}) = \mathrm{MaxPool} \left[\sigma \left(
    \textbf{H}^{(t-1)}   \textbf{W}_1
    \textbf{W}_2^{\top}\textbf{S}^{\top}
     \right)\right]\\\notag
&    V^{(t)} = \left\{w_n | P(w_n) > 0.5 \right\}
\end{align}
where $\textbf{W} \in \mathbb{R}^{d \times d}$ is a linear transformation matrix. This operation is similar to a multi-label classification. Every source word is assigned with an independent probability, and words with probabilities larger than $0.5$ are selected as new nodes $V^{(t)}$. To represent these new nodes as $\textbf{D}^{(t)}$ when their network structural information are unknown, we suppose that there are implicit edges pointing from previous nodes to these nodes. Besides, these new nodes are connected to each other by implicit edges. Although it is impossible to appear explicit edges among them, this operation can further enrich node representations. Their connections are illustrated by the second adjacency matrix in the middle block of Figure \ref{fig:flow}. Explicit edge connections and types are then figured out by Deep Biaffine Attention \cite{dozat2017deep}:
\begin{equation*}
    E^{(t)} = \mathrm{DeepBiaffine} \left(\Vert_{j=0}^{t-1} \textbf{H}^{(j)}, \textbf{D}^{(t)}\right)
\end{equation*}
where $\Vert_{j=0}^{t-1} \textbf{H}^{(j)}$ is achieved by concatenating head representations of all nodes in the previous hierarchies. The generation proceeds via repeating the aforementioned operations until no words can be selected as new nodes.

\subsection{Graph Representation Model}
Recently, Transformer \cite{vaswani2017attention} has made impressive progress in the graph representation field \cite{ying2021do}. In essence, Transformer regards inputs as an undirected fully-connected graph, thus serving as a special graph representation model that can enjoys global perception at all layers. Previous works focusing on adapting Transformer-encoder to node or graph classification, while this paper modifies Transformer-decoder to conduct graph generation. 

Let $\mathbf{x}_i^{(l)}$ denote the node $v_i$ embedding at the $l$-th layer. If the node $v_i$ is in the component $C_t$ and copied from the source word $w_n$, its initial node embedding $\mathbf{x}_i^{(0)}$ should be the summation of:
\begin{equation*}
    \mathbf{x}_i^{(0)} = \mathbf{S}_{[n]} + \mathbf{P}_{[t]}
\end{equation*}
where $\mathbf{S}, \mathbf{P}$ indicate word embeddings and hierarchical  positional encodings respectively, as shown in Figure \ref{fig:flow}. Nodes in the same hierarchy have the same hierarchical positional encodings. 

The message passing layer actually takes the position of the masked self-attention layer in the decoder. The original decoder self-attention helps every word to aggregate left-ward contexts. In contrast, every node in our model not only aggregates left-ward contexts (i.e, nodes in previous hierarchies), but also nodes in the same hierarchy. To distinguish explicit edges and implicit edges, the message vector $\mathbf{m}_{ji}$ of the node $v_j$ with an explicit edge pointing to the node $v_i$ should be enriched with prior structural knowledge by
\begin{equation*}
\mathbf{m}^{(l)}_{ji} =
    \begin{cases}
     \mathbf{x}^{(l)}_j + \mathrm{relu}\left(\mathbf{x}^{(l)}_j \mathbf{U}_{z_{ji}}\right)  , &  v_j \in \mathcal{N}_i \\
   \mathbf{x}^{(l)}_j ,& v_j \in \mathcal{D}_i
    \end{cases}
\end{equation*}
where $\mathbf{U}_{z_{ji}} \in \mathbb{R}^{d \times d}$ indicates the parametric embedding matrix of the edge label $z_{ji}$. These edge embedding metrics are shared across all layers. 
Notably, we assume that the central node $v_i$ is self-connected implicitly, i.e. $v_i \in \mathcal{D}_i$. The reduction function is then defined as the multi-head attention:
\begin{equation*}
    \alpha_{ji} = \frac{\exp \left(\mathbf{x}_i\mathbf{W}_Q \mathbf{W}_K^{\top}\mathbf{m}_{ji}^{\top} \right)}{\sum_{v_u \in \mathcal{N}_i \cup \mathcal{D}_i} \exp \left(\mathbf{x}_i\mathbf{W}_Q \mathbf{W}_K^{\top}\mathbf{m}_{ui}^{\top} \right)}
\end{equation*}

\begin{equation*}
    \mathbf{x}_i^{(l)'}= \left[\mathop{\Vert}\limits_{h=1}^{H} \left(\sum_{v_j \in \mathcal{N}_i \cup \mathcal{D}_i} \alpha_{ji}^h \mathbf{m}_{ji}^{(l)} \mathbf{W}_V^h \right)\right] \mathbf{W}_O
\end{equation*}
We can see that the query is the node embedding $\mathbf{x}_i$, and the keys and values are those message vectors $\mathbf{m}_{ji}$. Its output $\mathbf{x}_i^{(l)'}$ is then fed into the feed-forward layer to enter the next layer:
\begin{equation*}
    \mathbf{x}_{i}^{(l+1)} = \mathrm{FFN}\left( \mathbf{x}_i^{(l)'} \right)
\end{equation*}
The outputs $\mathbf{x}^{(L)}$ of the final layer are head representations or dependent representations.



\begin{table*}[htbp]\scriptsize
\centering
    \begin{tabular}{clcccccccccccccccc}
    \toprule
     &{\bf IWPT 2021 }&{\bf bg} & {\bf cs} & {\bf en}  & {\bf et} & {\bf fi} & {\bf fr}  & {\bf it} & {\bf lt} & {\bf lv}  & {\bf nl} &{\bf  pl} & {\bf ru}  & {\bf sk} & {\bf sv}  & {\bf uk} & {\bf avg}\\
   
    \midrule
     \multirow{3}{*}{\STAB{\rotatebox[origin=c]{90}{\it ELAS}}}
     &{\bf BiAtt}  & 92.7 & 91.0 & 87.2 & 87.2 & 90.6 &88.4 & 92.1 & 81.9 & 88.3& 90.5& 90.2 &93.2 & 91.5 & 87.3  &89.1 &89.4  \\
   
     &{\bf Tree-G} & 92.8 & {\bf 91.1} & 87.3 & 87.1 & {\bf 90.7} &88.6 & 92.3&81.9 & 88.2 & 90.5& {\bf 90.4} & 93.2 &91.6 & 87.5 & 89.0 &89.5  \\

      &{\bf SAGER} & {\bf 92.9} & 90.9 & {\bf 87.9} & {\bf 87.3} & {\bf 90.7} &{\bf 89.5} & {\bf 92.8} & {\bf 83.5} & {\bf 88.5}& {\bf 90.9} & {\bf 90.4} &{\bf 93.5} & {\bf 92.1} & {\bf 87.9}  &{\bf 89.6} & {\bf 89.9} \\

    \midrule
    \multirow{3}{*}{\STAB{\rotatebox[origin=c]{90}{\it GMS}}}
    &{\bf BiAtt}  & 47.4 & 44.8 & 36.3 & 37.1 & 38.7 &40.3 & 43.8 & 21.0 & 38.4& 46.9& 40.8 &50.3 & 51.0 & 32.2  &34.6 &40.2  \\
   
     &{\bf Tree-G} & 47.8 & 45.3 & 36.9 & 37.0 & 39.1 &41.1 & 44.7&21.0 & 37.9 & 47.0& 41.9 & 50.8 &51.5 & 33.4 & 34.2 &40.6  \\
      &{\bf SAGER} & {\bf 48.8 }& {\bf 45.6} & {\bf 40.3 }& {\bf 39.2} & {\bf 41.4} &{\bf 45.4} & {\bf 47.1} & {\bf 28.2} & {\bf 42.8}& {\bf 51.3} & {\bf 43.6} &{\bf 54.2 }& {\bf 57.8} & {\bf 38.2 } &{\bf 39.2} & {\bf 44.2} \\
     \midrule
     \midrule
     &{\bf IWPT 2020 }&{\bf bg} & {\bf cs} & {\bf en}  & {\bf et} & {\bf fi} & {\bf fr}  & {\bf it} & {\bf lt} & {\bf lv}  & {\bf nl} &{\bf  pl} & {\bf ru}  & {\bf sk} & {\bf sv}  & {\bf uk} & {\bf avg}\\
 
    \midrule
     \multirow{3}{*}{\STAB{\rotatebox[origin=c]{90}{\it ELAS}}}
     &{\bf Sec-order}   & 91.5 & 90.1 & 87.1 &86.0 & 89.0 & 85.3 & 91.5 & 78.9  & 87.6& 86.2 & 84.0 & 92.3 & 87.6 & 84.7 & 88.0 & 87.3 \\
   
     &{\bf UDify }  & 90.7 & 87.5 & 87.2 &84.5 & 89.5 & 85.9 & 91.5 & 77.6  & 84.9& 84.7 & 84.6 & 90.7 & 88.6 & 85.6 & 87.2 & 86.7 \\

      & {\bf SAGER} & {\bf 92.6} & {\bf 90.4} & {\bf 88.2} & {\bf 86.9} & {\bf 90.1} & {\bf 87.4} & {\bf 92.6} & {\bf 82.5} & {\bf 88.5}& {\bf 86.7} &{\bf 86.7} &{\bf 93.2} & {\bf 91.0} & {\bf 87.0} & {\bf 89.0 } & {\bf 88.9 }\\

    \midrule
    \multirow{3}{*}{\STAB{\rotatebox[origin=c]{90}{\it GMS}}}
    &{\bf Sec-order}  & 43.1 & 37.7 & 35.7 & 31.8 & 34.4 &29.2 & 44.4 & 15.1 & 35.3& 31.0& 28.6 &47.1 & 38.7 & 26.5  &30.5 &33.9  \\
   
     &{\bf UDify }  & 41.4 & 31.4 & 34.1 &31.2 & 34.5 & 33.2 & 41.5 & 17.8  & 31.6& 23.8 & 26.1 & 40.6 & 43.1 & 27.1 & 31.2 & 32.6 \\
      &{\bf SAGER} & {\bf 48.3} & {\bf 43.5} &{\bf  41.6} & {\bf 36.6} & {\bf 38.4} &{\bf 38.7} & {\bf 47.1} & {\bf 25.6} & {\bf 42.0}&{\bf 34.1 }& {\bf 32.9}&{\bf 53.7}& {\bf 55.2} &{\bf  36.4}  &{\bf 38.3} & {\bf40.8} \\
     \bottomrule
  \end{tabular}
   \caption{Average ELAS and GMS results of $3$ calculations on IWPT 2021 and IWPT 2020 datasets. We use $L=2$ according to ELAS on the English dev-set. }
     \label{tab:main}
\end{table*}

The edge embedding matrices $\mathbf{U}$ give the model access to prior structural knowledge and enable it to select useful prior knowledge adaptively. When all structural knowledge is useless (i.e, parameters in $\mathbf{U}$ are trained to be zeros) and each hierarchy only contains one node, the graph model degrades to a vanilla Transformer decoder.

\section{Experiment}
{\bf Datasets}. 
We tune our models primarily on $15$ languages that appear in IWPT 2020 dataset and IWPT 2021 dataset  \cite{bouma2020overview, bouma2021raw}. The two shared tasks focus on EUD \cite{schuster2016enhanced} which are non-tree graphs with reentrancies, empty nodes and sparsity cycles. To construct the topological hierarchy, we need to delete the back edges in cycles firstly and add them back by rules at inference time. For the language that has multiple treebanks, we simply concatenate all of its treebanks. Besides, we use gold tokenization and gold sentence segmentation during training and development. At test time, we use the results of tokenization and segmentation provided by the top ranked models. 


\vspace{0.1in}
\noindent{\bf Baseline Models}.
Our comparison experiments aim to investigate the performances of models themselves, without considering some learning techniques like ensembling \cite{grunewald2021simple}, two-stage training \cite{shi2021tgif} and automated concatenation of embeddings \cite{wang2021enhanced}. We conclude four strong baselines from top-ranked systems in IWPT 2021 and IWPT 2020, namely Deep Biaffine Attention \cite{dozat2017deep}, Tree-Graph Parser \cite{shi2021tgif}, Second-order Parser \cite{wang2019second, wang2020enhanced} and Language-specific UDify \cite{kondratyuk201975, kanerva2020turku}. Their results are reported after eliminating the effects of learning techniques.


\vspace{0.1in}
\noindent{\bf Evaluation Metrics}.
ELAS results are evaluated by the official script provided by IWPT 2021. Besides, we also define a graph-level matching score (GMS) to investigate whether the model can deal with a few arcs that are difficult to be predicted properly in a sample. Since we segment UD sentences from raw texts, the numbers of sentences are different for each system. Therefore, GMS is an $F_1$ score around the number of absolutely matched graphs,
\begin{equation*}
    Recall = \frac{\#matched\ graph}{\# gold\ sentences}
\end{equation*}
\begin{equation*}
    Precision = \frac{\#matched\ graph}{\# system\ sentences}
\end{equation*}

\begin{equation*}
   GMS = \frac{2 \times  Recall \times Precision}{Recall + Precision}
\end{equation*}
where $\#$ represents $The\ number\ of$.

\vspace{0.1in}
\noindent{\bf Word Embeddings}.
Similar to the operations in most top-ranked systems, our word embeddings $\mathbf{S}_{[n]}$ are initialized as the weighted summation of the corresponding hidden states in XLM-R layers \cite{conneau2020unsupervised}, where the weights are the learned attention distribution over all XLM-R layers. For the word composed of multiple subwords, we extract the hidden states of the last one. We set up the dimension in the graph representation model as $d = 1024$, the same as that in the pre-trained models. 

\vspace{0.1in}
\noindent{\bf Other Details}.
We train models directly on each language with Teacher Forcing Training to output all head (dependent) representations at once. Besides, we truncate the input sentences to $100$ words at training time. We totally run $100$ epochs with $16$ batch size and select the model parameters base on the ELAS on the development sets.  We train our models on a single v100 with a speed of about $10000$ samples in $10$ minutes. We use ReZero \cite{bachlechner2021rezero} in our graph transformer, instead of LayerNorm operations commonly used in Transformer. In this case, we do not need to use the warm-up learning schedule, and we use Adam optimizer with the $0.97$ decay ratio of the learning rate. We set up the initial learning rate of pre-trained embeddings as $2e-5$, and that of others as $1e-3$. Besides, dropout rates in the part of pre-training and graph representation are set to $0.1$, while the output layers of nodes and edges are set to $0.3$.  We build up the vocabulary on the arc labels 
for each language respectively. To shrink the size of edge vocabularies, we follow the de-lexicalization operations of arc labels \cite{grunewald2020surprisingly} and re-lexicalize them before evaluations.



\begin{figure*}
\centering
\includegraphics[width=0.95\textwidth]{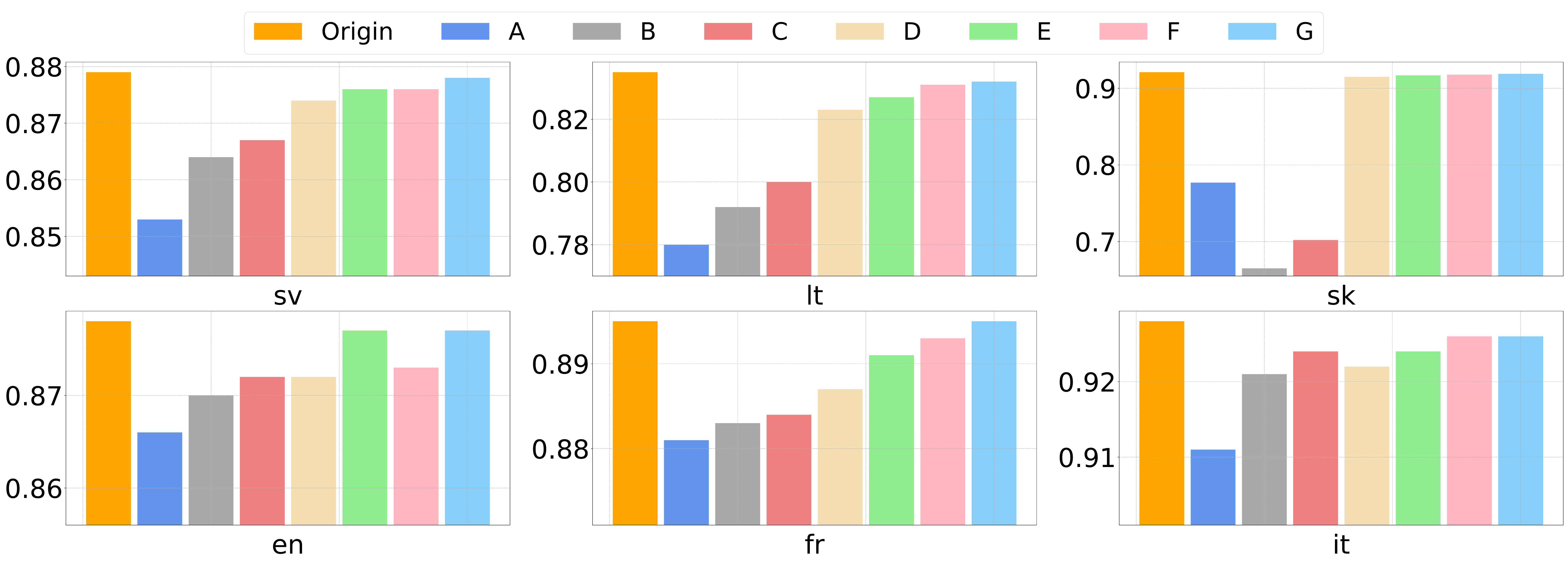}
\caption{\label{fig:variant} Test-set ELAS results, comparing the origin model with different model variations}
\end{figure*}

\section{Results and Analysis}
\subsection{Main Results}
The official evaluation metrics ELAS of our models and baselines are shown in Table \ref{tab:main}. We note that SAGER achieves at least comparable results on all languages. In IWPT 2021, in addition to obtaining the best average ELAS performance (average $\sim 0.4\%$ points), our model brings significant improvements over multiple languages like Lithuanian ($\sim 1.6\%$ points), French ($\sim 0.9\%$ points), English ($\sim 0.6\%$ points). This enhancement is more significant when comparing our model with the top two models in IWPT 2020 (average $\sim 1.6\%$ points ). Besides, sharper increases appears in GMS of IWPT 2021 (average $\sim 3.6\%$ points) and IWPT 2020 (average $\sim 6.9\%$ points), where our model achieves an amazing rising against the baselines in all languages. 

It should be mentioned that a higher ELAS does not mean a higher GMS, as shown in the results of Czech ({\it cs}) language. In other words, some obstinate errors are fixed to make more dependency graphs completely correct, but there appear some samples where more mistakes concentrate. This situation derives from the inherent characteristics of autoregressive generation that the prediction accuracy at one certain step is heavily dependent on that at historical steps. In ideal states, the historical information can calibrate some obstinate errors by the learned dependencies. However, once deviation occurs in an intermediate step, it may lead to mistakes that should not have been made. This is the essential reason that autoregressive parsers are weaker than non-autoregressive parsers. By comparison, our semi-autoregressive parser mitigates the negative impact of this characteristic by removing some unreasonable dependency relationships, thus resulting in better performances in both ELAS and GMS.

\subsection{Model Variant Ablation Studies}
\label{sec:ab}
To investigate the importance of different model components and
input features, we evaluated the following variations of our model.

{\bf A. Autoregressive generation with random orders}. We impose random orders to the sibling nodes, so the model is converted to a fully-autoregressive generator. At each step, the model only generates a new node and its all incoming edges. The sibling nodes will be re-ordered after a training epoch.

{\bf B. Autoregressive generation with word orders}. The sibling nodes are sorted by the  positions of the node words in the sentence.

{\bf C. Combine random orders and word orders}. The sibling nodes are firstly randomly sorted at the early stage of training and fixed to the word orders at later training.

{\bf D. No implicit edges}. Without the implicit edges, the graph representation model is similar to GAT \cite{velivckovic2018graph} but the messages are additionally enriched with the arc label information.

{\bf E. No implicit edges in the same hierarchy}. We remove the implicit edges between nodes in the same topological hierarchy. In this case, each node only has the incoming arcs from the nodes in the previous hierarchies.

{\bf F. No explicit edges}. We replace all explicit edges by implicit edges, which is equal to forcing the edge embedding matrix $\mathbf{U}$ to zeros.

{\bf G. No hierarchical positional encodings}. In this case, the model would lose the sequential relationships between hierarchies and fail to locate nodes of different hierarchy.

The ablation results of $6$ languages are summarized in Figure \ref{fig:variant}. We firstly focus on the fully-autoregressive variations, namely the model A, B and C. We can see that there are significant declines in performances when imposing orderings to sibling nodes, indicating that the autoregressive mode heavily suffers from exposure bias in terms of generation orderings. Besides, the extent of declines varies a lot in different languages, ranging from over $30\%$ in the Slovak ({\it sk}) dataset and within $1\%$ in the English ({\it en}) and Italian ({\it it}) datasets. This proves that the impact of imposed sorting is quite unstable. 

Moving to the model D, E and F which are variations with respect to explicit and implicit edges. Although it is not as significant as the negative effects of using autoregressive modes, that of removing implicit or explicit edges cannot be ignored. Generally, implicit edges play a more important role than explicit edges as the performances of model D are often lower than those of model E and F. This validates that the edge sparsity is the major problem of graph generation after the uncertainty of generation orderings. Besides, the implicit edges in the same hierarchy (see model E) and the historical arc label information (see model F) are both compulsory model components because the model always performs worse when dropping them. The final is the model without hierarchical positional encodings. Compared with other variations, its performance is the closest to the origin model, implying that our graph representation model is not very sensitive to the sequential relationships between hierarchies.

\subsection{Error analysis of Topological Hierarchy}

\begin{figure}[h!]
\centering
\includegraphics[width=0.48\textwidth]{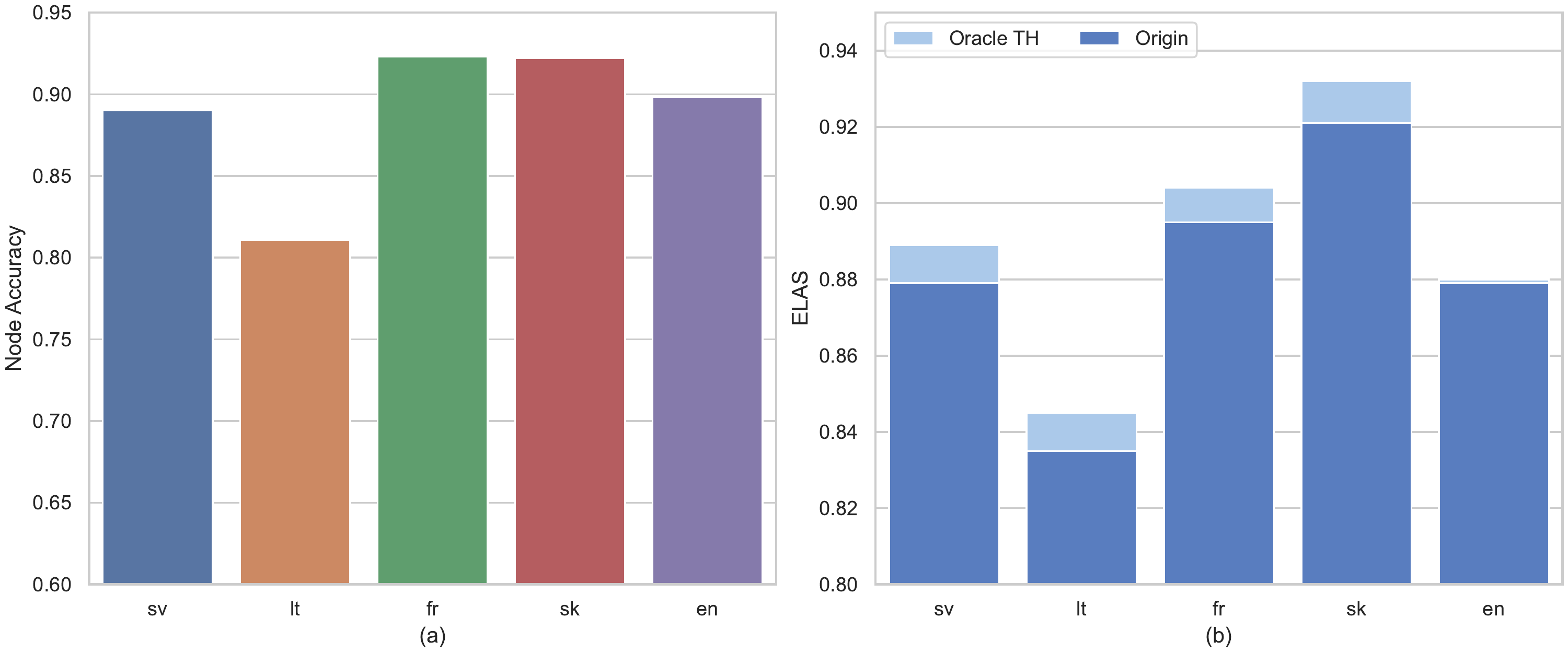}
\caption{\label{fig:error} (a) Accuracy of nodes in the correct hierarchy. (b) ELAS results using oracle Topological Hierarchy (TH). }
\end{figure}

Since a topological hierarchy regulates the rough topological structure of a dependency graph, its prediction accuracy is crucial for the whole model. We investigate the node accuracy on $5$ languages (see Figure \ref{fig:error}.a), and find that about $90\%$ nodes can fall into correct hierarchies. Even the language performing worst under the ELAS evaluation can reach $80\%$ node accuracy. We then provide the model with the oracle node group at each generation step and plot the comparison results against the origin model in Figure \ref{fig:error}.b. There is about a $1\%$ increase of ELAS on most languages when using oracle TH. It is surprising that oracle TH does not bring about improvement to the English dataset, indicating that corrupt topological hierarchies do not always lead to incorrect arcs. Actually, it would cause bad results only when the dependent node of an arc is generated before its head nodes. It does not matter for corruptions that do not shuffle the orders of heads and dependents.


\subsection{Sensitive analysis of Layers}

\begin{figure}[h]
\centering
\includegraphics[width=0.48\textwidth]{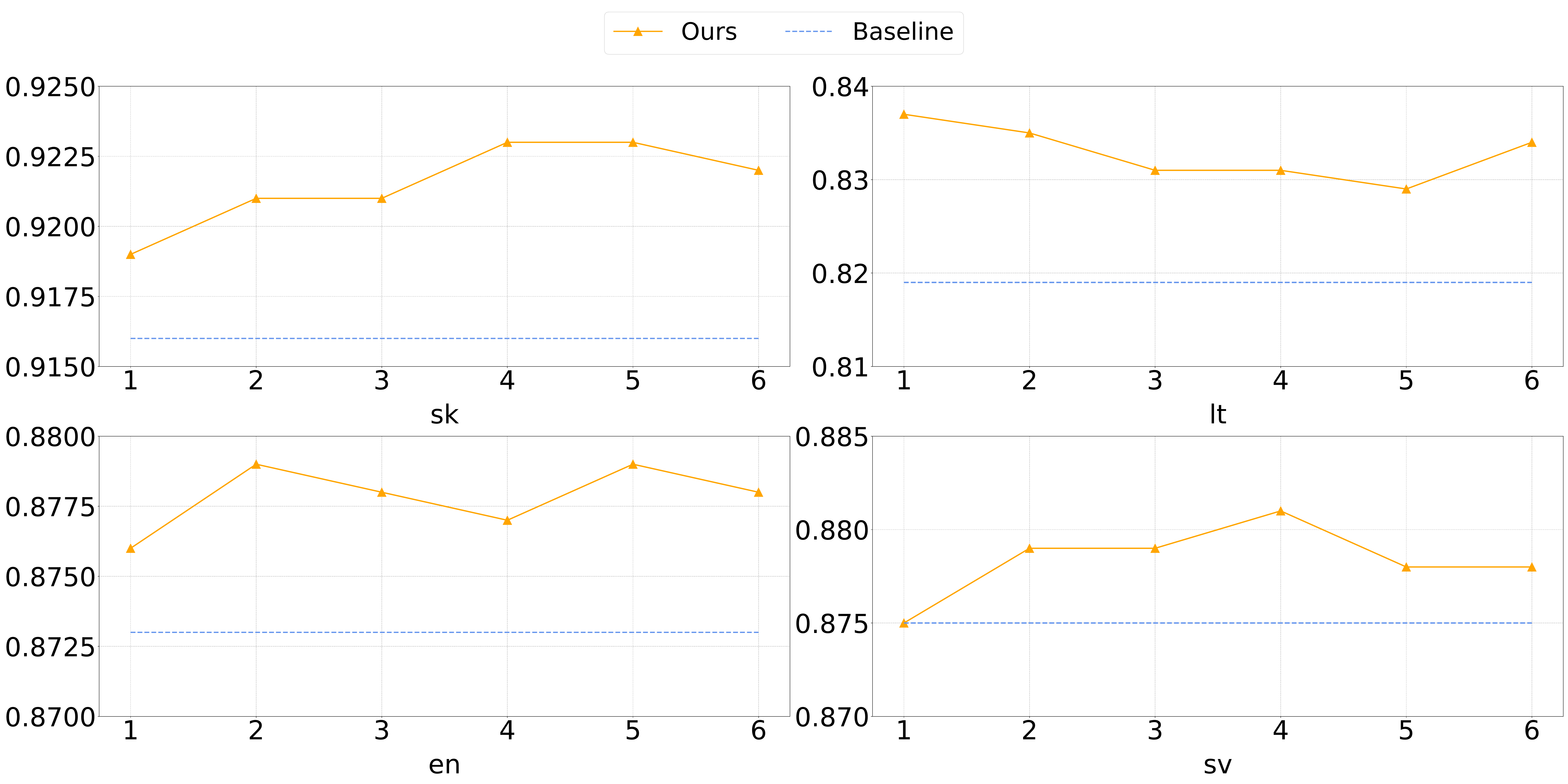}
\caption{\label{fig:layer} Sensitive analysis of layers on test sets. }
\end{figure}

We test the sensitivity of ELAS results to different $L$. As shown in Figure \ref{fig:layer}, We select four languages whose ELAS are significantly higher than baselines' when $L=2$. We find that our model still outperforms the baseline whichever $L$ is used. Besides, it is hard to disclose a trend between model performance and the number of layers from the four plots. This is possibly because graph transformer can capture context information of high-order neighbours even with one layer. Overall, SAGER is insensitive to the number of layers.

\section{Case Study}

\begin{figure}[h]
\centering
\includegraphics[width=0.48\textwidth]{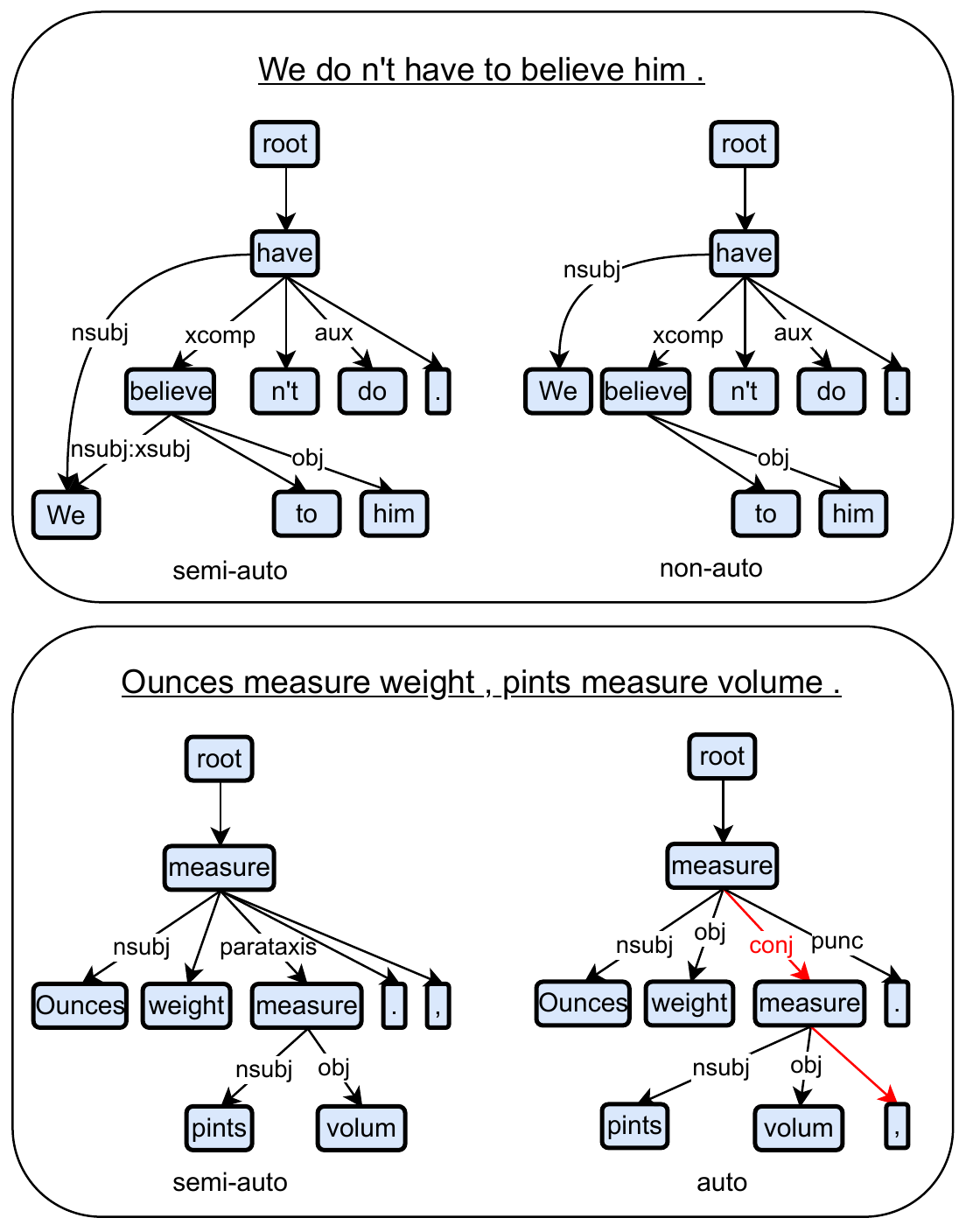}
\caption{\label{fig:case} Case study, auto is the model B in \cref{sec:ab}.}
\end{figure}

To better understand the drawbacks of non-autoregressive and autoregressive models, we present two examples in Figure \ref{fig:case}. In the first example, semi-auto performs better because it can predict the arc {\it nsubj:xsubj} with the help of the previously predicted arc {\it xcomp}. The two edges are heavily dependent. In contrast, non-auto fails because it assumes independency between edges. In the graph produced by the auto, the edge between the first {\it measure} and the comma should have been predicted after linking {\it measure} to {\it weight}, but the model actually skips the step and predicts a wrong arc {\it conj}. This is probably because {\it (nsubj, obj, conj)} is a very highly-frequent dependent relationship in autoregressive graph sequences.

\section{Conclusion and Limitation}
This paper explores a semi-autoregressive dependency graph parser (SAGER) that learns the explicit dependencies in dependency graphs. This generation pattern captures the edge dependencies while reducing exposure bias, resulting in a more effective parser and two insights about graph generation, namely the ordering uncertainty and edge sparsity. We believe that the proposed algorithm can be applied in other language parsing problems such as OIA (Open Information Annotation)~\citep{sun2020predicate,wang2022oie} and graph generation problems~\citep{sun2019graph}. 

The limitations of our work are inference speed and decoding strategy. The efficiency of semi-autoregressive inference is lower than that of non-autoregressive algorithms, so currently it cannot  be applied to scenarios with high frequency requests. Furthermore, this paper only introduces greedy search as a decoding strategy. Overcoming the challenge of introducing many complex decoding strategies such as beam/tree search~\cite{liu2020extracting, ma2021global} belongs to our future work.


\bibliography{refs_scholar}
\bibliographystyle{acl_natbib}

\end{document}